\documentclass[11pt]{article}

\usepackage[final]{acl}

\usepackage{times}
\usepackage{latexsym}
\usepackage{amssymb}
\usepackage{amsmath}
\usepackage{booktabs} 
\usepackage{multirow}
\usepackage[skins,breakable]{tcolorbox}
\usepackage[T1]{fontenc}
\usepackage[utf8]{inputenc}

\usepackage{microtype}

\usepackage{inconsolata}

\usepackage{graphicx}
\usepackage[table]{xcolor}
\title{\textsc{MAVEN}: Evidence-State Rewards for Long-Context Reasoning}

\author{
 \textbf{Ya Gao\textsuperscript{1}},
 \textbf{Pekka Marttinen\textsuperscript{1}}
\\
\\
 \textsuperscript{1}Aalto University
\\
 \small{
   \textbf{Correspondence:} \href{mailto:ya.gao@aalto.fi}{ya.gao@aalto.fi}
 }
}

\begin{document}
\maketitle
\begin{abstract}
Long-context reasoning requires models to locate, revise, and synthesize evidence distributed across lengthy inputs. Existing long-context RL methods usually reward final answers or static evidence extraction, offering little feedback on how intermediate actions change the model's evidence state. We propose \textsc{Maven} (\textbf{Ma}rginal-\textbf{V}alue \textbf{E}vidence \textbf{N}avigation), a reinforcement learning framework with an editable evidence memory. \textsc{Maven} defines an answer-conditioned evidence-state value and rewards action-level state transitions: add actions are credited by marginal gain and hindsight contribution, link actions by evidence synergy, and drop actions by improved answer support after removing misleading evidence. These rewards are assigned to the corresponding action spans in GRPO. Across Llama and Qwen models on LongBench v2, LongReason, and RULER, \textsc{Maven} outperforms outcome-only RL and evidence-identification baselines, producing more sufficient evidence sets and lower distractor retention. Our results show that long-context RL benefits from optimizing stateful evidence navigation rather than one-shot evidence extraction.
\end{abstract}

\section{Introduction}

\begin{figure}[t]
    \centering
    \includegraphics[width=1.0\linewidth]{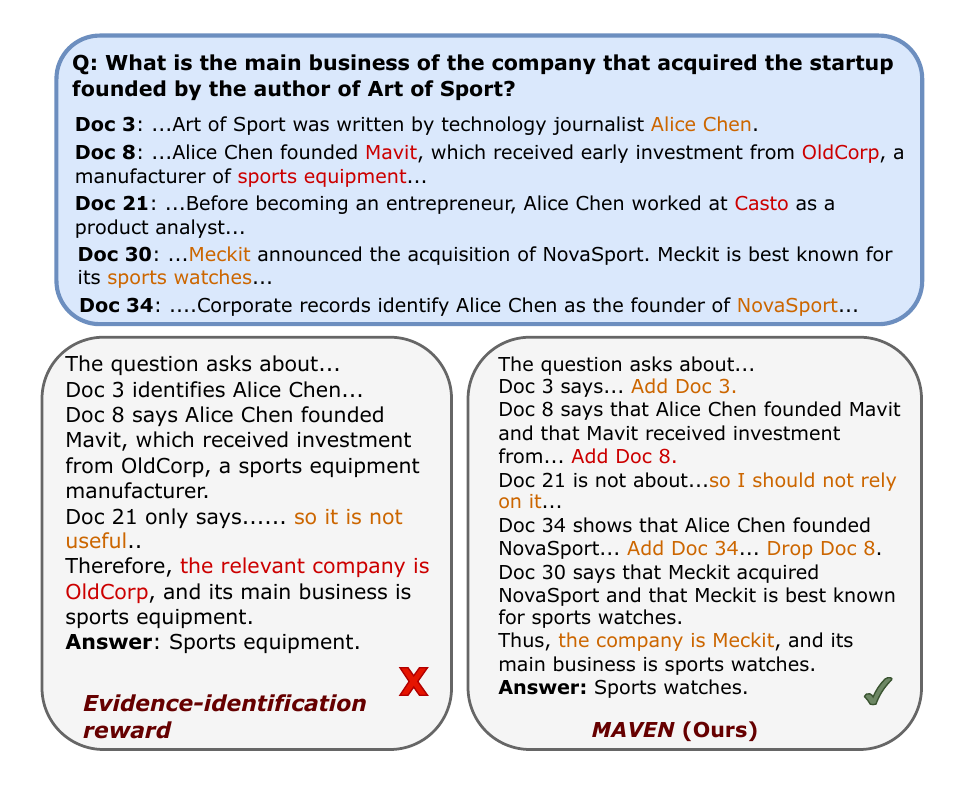}
    \caption{Motivating example illustrating the limitation of evidence-identification rewards. Although the baseline finds a relevant first hop, it anchors on a plausible distractor and stops before recovering the full evidence chain. In contrast, \textsc{Maven} encourages exploration, where the model revises its evidence state by dropping the misleading document, adding the missing evidence, and producing the correct answer.
    }
    \label{fig:intro}
\end{figure}

Advanced Large Language Models (LLMs), especially reasoning-oriented models, have shown strong performance on complex tasks such as mathematical reasoning, coding, and multi-step problem solving~\citep{jaech2024openai,guo2025deepseek,team2025kimiagent}. As LLMs are increasingly used in real-world workflows, they are also expected to process and reason over large volumes of information. Recent models therefore support increasingly long context windows~\citep{team2025kimi}. However, the ability to accept long inputs does not necessarily imply the ability to reason effectively over them. Long-context models may still suffer from issues such as ``lost in the middle''~\citep{liu2024lost}, premature reliance on partial evidence, failure to identify necessary information, or inability to synthesize evidence distributed across distant parts of the context. 

Reinforcement Learning (RL) has become an effective approach for improving the reasoning behavior of LLMs~\citep{jin2025search}, and recent work has begun to apply RL to long-context reasoning. A straightforward strategy is to optimize the model using only outcome rewards, which is based solely on verification of the final answer~\citep{wan2025qwenlong,wang2025loongrl}. While simple and scalable, outcome-only rewards provide limited feedback about the intermediate process~\citep{chen2026longrlvr}. Beyond outcome-only supervision, recent studies introduce context-aware or evidence-aware rewards~\citep{chen2026longrlvr,guan2026evidence,ping2026longr}. These rewards commonly supervise evidence-related behavior, such as selecting gold chunks, quoting context segments, or assessing the quality of extracted evidence. Such designs improve over purely answer-level feedback. However, they largely centered on evidence identification or evidence quality, where they evaluate evidence as isolated chunks, quoted spans, or a final extracted set. As a result, these rewards provide useful guidance for retrieval, but they do not fully capture the dynamic and state-dependent nature of long-context reasoning.

In challenging long-context tasks, evidence is rarely useful in isolation. A segment may become valuable only after another segment is found; a plausible document may need to be discarded after later evidence reveals it as a distractor; and two individually insufficient pieces may jointly bridge a reasoning gap. As shown in Figure~\ref{fig:intro}, an extraction-centric reward can encourage premature commitment: once the model finds a locally plausible chunk, it may stop exploring, anchor on an incorrect evidence path, or fail to revise its evidence set. Effective long-context reasoning therefore requires more than retrieving relevant chunks. The model should learn to build, revise, and synthesize an evolving evidence state.

We propose \textsc{Maven} (\textbf{Ma}rginal-\textbf{V}alue \textbf{E}vidence \textbf{N}avigation), a reinforcement learning framework with an editable evidence memory. The model learns to add evidence, link complementary evidence, drop misleading evidence, and answer the question. Rather than rewarding isolated evidence extraction, \textsc{Maven} scores how each action changes the current answer-supporting evidence state.  
% In this way, \textsc{Maven} trains the model to navigate long contexts as a process of evidence-state construction rather than one-shot evidence extraction.

We evaluate \textsc{Maven} on Llama-3.1-8B, Qwen2.5-14B, and Qwen3-30B-A3B across LongBench v2, LongReason, and RULER. \textsc{Maven} consistently improves final-answer performance, while also increasing evidence sufficiency and reducing distractor retention. 
% Further analyses demonstrate that the proposed add, link, and drop rewards contribute complementary benefits, and that action-local process supervision improves the model's ability to revise and synthesize evidence during long-context reasoning.
Our contributions are:
\begin{itemize}
    \item We identify a limitation of existing long-context RL rewards: they often supervise static evidence extraction rather than evidence-state transitions.
    \item We propose \textsc{Maven}, an editable evidence-memory framework that defines an answer-conditioned evidence-state value and trains LLMs to add, link, drop, and answer through stateful, action-local process rewards.
    \item We show consistent gains across model families and benchmarks, with diagnostic evidence that \textsc{Maven} improves sufficiency and reduces distractor retention.
\end{itemize}
\section{Methodology}
\label{sec:method}

\begin{figure*}[t]
    \centering
    \includegraphics[width=1.1\textwidth]{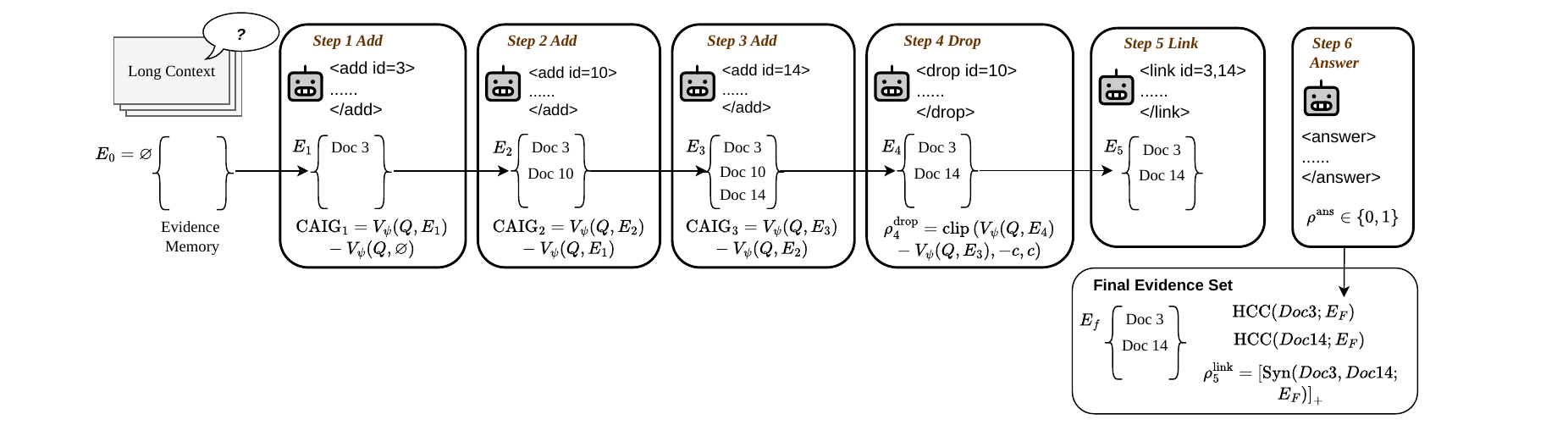}
    \caption{The model edits an evidence memory through \texttt{add}, \texttt{drop}, and \texttt{link} actions before answering. Actions receive local process rewards based on their effect on the evidence state, and final kept evidence receives hindsight credit for its contribution to the final evidence set.}
    \label{fig:method}
\end{figure*}

% We propose \textsc{Maven}, a reinforcement learning framework for long-context reasoning with an editable evidence memory. Instead of rewarding isolated evidence extraction, \textsc{Maven} rewards how each generated action changes the model's evidence state with respect to the final answer. The model can add evidence, link evidence pieces, drop misleading evidence, and produce the final answer. 
\textsc{Maven} trains a policy to edit an evidence memory before answering. Each generated action changes, explains, or uses the current memory, and receives a reward based on its contribution to answer support.

\subsection{Problem Formulation}
\label{sec:problem_formulation}
Each training instance is denoted as: $x = (C, Q, a^\star)$, where \(C\) is a long context, \(Q\) is the question, and \(a^\star\) is the gold answer. The policy model \(\pi_\theta\) receives \((C,Q)\) as input and generates a trajectory which is parsed into a sequence of high-level actions. Each action \(m_k\) corresponds to a contiguous span of generated tokens \(I_k \subseteq \{1,\ldots,T\}\).
The model is allowed to use four action types: (1) \texttt{<add id=i> ... </add>}: add a quoted or referenced evidence span with the source id $i$ to the evidence memory; (2) \texttt{<link ids=i,j> ... </link>}: explain how two evidence pieces jointly support the answer; (3) \texttt{<drop id=i> ... </drop>}: remove an evidence piece that is irrelevant, redundant, or misleading; (4) \texttt{<answer> ... </answer>}: produce the final answer. The action order is not predetermined. The model may add, link, and drop evidence multiple times before answering. 

Let $E_k=\{e_1, e_2, \ldots, e_{n_k}\}$, a set of evidence items, denote the evidence memory after the \(k\)-th action, with \(E_0=\varnothing\). Add and drop actions modify the memory, while link actions leave the memory unchanged. The final kept evidence set is denoted by \(E_F\).
% , and the final valid link set is denoted by \(L_F\).

We impose only lightweight structural constraints: each evidence item must have a unique identifier, drop and link actions must refer to existing identifiers, the number of kept evidence items is capped by \(K_{\max}\), and evidence spans exceeding a maximum length are invalid. These constraints control computational cost and prevent degenerate add--drop loops.

\subsection{Answer-Conditioned Evidence-State Value}
\label{sec:evidence_value}

The central component of \textsc{Maven} is an evidence-state value function that measures whether the current evidence memory helps predict the gold answer. We use a frozen verifier model \(p_\psi\), rather than training a separate reward model in the main method.

Given a question \(Q\), evidence memory \(E\), and gold answer \(a^\star=(a^\star_1,\ldots,a^\star_{|a^\star|})\), the verifier computes the teacher-forced negative log-likelihood:
\begin{equation}
    \ell_\psi(Q,E)
    =
    -\frac{1}{|a^\star|}
    \sum_{j=1}^{|a^\star|}
    \log p_\psi
    \left(
        a^\star_j
        \mid
        Q, E, a^\star_{<j}
    \right).
\end{equation}
We then define the normalized evidence-state value:
\begin{equation}
    V_\psi(Q,E)
    =\frac{
        \ell_\psi(Q,\varnothing) - \ell_\psi(Q,E)}{
        \ell_\psi(Q,\varnothing) + \epsilon}.
    \label{eq:value}
\end{equation}

This value measures the relative reduction in answer NLL caused by the evidence memory. A positive value means that the evidence makes the correct answer more predictable.

\subsection{Action Rewards}
\label{sec:action_rewards}
For each generated trajectory, \textsc{Maven} computes rewards for four action types: add, drop, link, and answer. These rewards are later assigned to the corresponding generated action spans.

\subsubsection{Add Reward}
\label{sec:add_reward}
When the model adds a new evidence item \(e_k\), the evidence memory changes from \(E_{k-1}\) to \(E_k\). We first compute the online Conditional Answer Information Gain:
\begin{equation}
\begin{split}
    \operatorname{CAIG}_k(e_k)
    =
    V_\psi(Q,E_k) - V_\psi(Q,E_{k-1}) \\
    =    \frac{
        \ell_\psi(Q,E_{k-1}) - \ell_\psi(Q,E_k)
    }{
        \ell_\psi(Q,\varnothing)+\epsilon
    }.
    \label{eq:caig}
\end{split}
\end{equation}
This term asks whether the newly added evidence improves the current evidence state, conditioned on what has already been collected.

However, online marginal gain alone can under-credit early evidence in multi-hop reasoning. Suppose two evidence pieces \(e_i\) and \(e_j\) are only useful together, which means: $ V_\psi(Q,\{e_i\}) \approx 0$, $V_\psi(Q,\{e_j\}) \approx 0$, and
$V_\psi(Q,\{e_i,e_j\}) > 0$.
If \(e_i\) is added first, it may receive little online reward even though it is necessary for the final reasoning chain.

To address this, we add a hindsight credit term. After the trajectory terminates, for each \emph{final kept} evidence item \(e_i\in E_F\), we compute:
\begin{equation}
\begin{split}
\operatorname{HCC}(e_i;E_F)=[V_\psi(Q,E_F) \\
- V_\psi(Q,E_F\setminus\{e_i\})]_+.
\end{split}
\label{eq:hcc}
\end{equation}
where \([z]_+=\max(z,0)\). This term measures how much the final evidence value would decrease if \(e_i\) were removed. It is an efficient leave-one-out approximation to Shapley-style evidence credit.

The reward for a valid add action is:
\begin{equation}
\begin{split}
    \rho_k^{\mathrm{add}}
    =
    \alpha \,
    \operatorname{clip}
    \left(
    \operatorname{CAIG}_k(e_k), -c, c
    \right) \\
    +
    (1-\alpha)
    \mathbf{1}[e_k\in E_F]
    \operatorname{HCC}(e_k;E_F).
    \label{eq:add_reward}
\end{split}
\end{equation}
where \(\alpha\in[0,1]\) balances online progress and hindsight credit.

Thus, an added item is rewarded either for immediately improving the current state or for being necessary in the final evidence memory. Items later dropped receive no hindsight credit.

\subsubsection{Drop Reward}
\label{sec:drop_reward}

The drop action allows the model to revise its evidence memory. If action \(m_k\) drops evidence \(e\in E_{k-1}\), then \(E_k=E_{k-1}\setminus\{e\}\). The drop reward is:
\begin{equation}
    \rho_k^{\mathrm{drop}}
    =
    \operatorname{clip}
    \left(
    V_\psi(Q,E_k) - V_\psi(Q,E_{k-1}),
    -c,c
    \right).
    \label{eq:drop_reward}
\end{equation}

This reward is positive when removing evidence improves the answer-conditioned evidence state and negative when removing evidence harms it. Therefore, the model can recover from adding distracting evidence, but dropping useful evidence is penalized.

\subsubsection{Link Reward}
\label{sec:link_reward}

Long-context reasoning often requires synthesis: two evidence pieces may be individually insufficient but jointly decisive. To encourage explicit synthesis, \textsc{Maven} rewards link actions between complementary evidence pieces.

We define the pairwise synergy score:
\begin{align}
    \operatorname{Syn}(e_i,e_j;E_F)
    &=
    V_\psi(Q,E_F)
    -
    V_\psi(Q,E_F\setminus\{e_i\})
    \nonumber\\
    &\quad
    -
    V_\psi(Q,E_F\setminus\{e_j\})
    \nonumber\\
    &\quad
    +
    V_\psi(Q,E_F\setminus\{e_i,e_j\}).
\end{align}
The link reward is:
\begin{equation}
    \rho_k^{\mathrm{link}}
    =
    \left[
    \operatorname{Syn}(e_i,e_j;E_F)
    \right]_+.
    \label{eq:link_reward}
\end{equation}

This term is positive when \(e_i\) and \(e_j\) are more valuable together than separately. If either evidence piece is not retained in the final evidence set, the synergy score is set to zero. If two linked evidence pieces are redundant, the synergy score is near zero or negative, and the link receives no positive reward. Multi-hop reasoning can be represented by multiple pairwise links, forming an evidence graph over \(E_F\).

\subsubsection{Answer Reward}
\label{sec:answer_reward}
We score the final answer using a substring match to measure whether the ground truth is covered by the final answer, and define a binary answer reward \(\rho^{\mathrm{ans}}\in\{0,1\}\). This keeps training aligned with task success while add, drop, and link rewards provide dense process supervision.

\subsection{Training Objective and Procedure}
\label{sec:rl_objective}

We optimize the policy with GRPO~\citep{shao2024deepseekmath}. For each input $x$, the old policy \(\pi_{\theta_{\mathrm{old}}}\) samples a group of \(G\) trajectories.
For the $k$-th action in trajectory $i$, we compute an action reward $\rho_{i,k} \in \{\rho_{i,k}^{\mathrm{add}}, \rho_{i,k}^{\mathrm{drop}}, \rho_{i,k}^{\mathrm{link}}, \rho_{i,k}^{\mathrm{ans}} \}$.

Process rewards are assigned to the token spans that generated the corresponding actions rather than collapsed into a single scalar trajectory reward. We compute group-relative advantages separately for each action type. 
\(b(i,k)\) denotes the type of the $k$-th action in trajectory $i$. For each type, we collect all rewards of that type from the rollout group, denoted as $\mathcal{R}_{x,b}$.
The normalized action-level advantage is:
\begin{equation}
    \hat A_{i,k}
    =
    \lambda_{b(i,k)}
    \cdot
    \frac{
        \rho_{i,k}
        -
        \operatorname{mean}(\mathcal{R}_{x,b(i,k)})
    }{
        \operatorname{std}(\mathcal{R}_{x,b(i,k)})
    }.
    \label{eq:group_action_advantage}
\end{equation}
where \(\lambda_b\) is the weight for action type \(b\).

For token \(y_{i,t}\), the token-level importance ratio is:
\begin{equation}
    r_{i,t}(\theta)
    =
    \frac{
        \pi_\theta(y_{i,t}\mid C,Q,y_{i,<t})
    }{
        \pi_{\theta_{\mathrm{old}}}(y_{i,t}\mid C,Q,y_{i,<t})
    }.
\end{equation}
All tokens in the $k$-th action of the trajectory $i$ receive the same advantage. The GRPO objective is:
\begin{align}
    \mathcal{J}(\theta)
    &=
    \mathbb{E}_{x,\{\tau_i\}_{i=1}^{G}}
    \Bigg[
    \frac{1}{\sum_{i=1}^{G}K_i}
    \sum_{i=1}^{G}
    \sum_{k=1}^{K_i}
    \frac{1}{|I_{i,k}|}
    \sum_{t\in I_{i,k}}
    \nonumber\\
    &\quad
    \min
    \Big(
        r_{i,t}(\theta)\hat A_{i,k},
        \operatorname{Clip\big(i,t\big)}
        \hat A_{i,k}
    \Big)
    \Bigg]
    \nonumber\\
    &\quad
    -
    \beta D_{\mathrm{KL}}.
    \label{eq:action_grpo}
\end{align}
where $\operatorname{Clip\big(i,t\big)}=\operatorname{clip}
        \big(
            r_{i,t}(\theta),
            1-\epsilon_{\mathrm{c}},
            1+\epsilon_{\mathrm{c}}
        \big)$ and $D_{\mathrm{KL}}$ is a KL penalty against the reference policy. We use a small \(\beta\) during training for stability. This objective preserves process-level credit assignment. For example, if a trajectory contains a bad add action followed by a useful drop action, the add span can receive a negative advantage while the drop span receives a positive advantage.

Training proceeds in two stages. We first perform a small supervised fine-tuning stage to teach the model the editable evidence-memory interface. The demonstrations contain valid \texttt{add}, \texttt{link}, \texttt{drop}, and \texttt{answer} actions, but do not impose a fixed action order. 
% The SFT loss is:
% \begin{equation}
%     \mathcal{L}_{\mathrm{SFT}}(\theta)
%     =
%     -
%     \mathbb{E}_{(x,\tau^\star)}
%     \sum_{t=1}^{|\tau^\star|}
%     \log
%     \pi_\theta
%     \left(
%         y_t^\star
%         \mid
%         C,Q,y^\star_{<t}
%     \right).
% \end{equation}
This stage is used only to reduce invalid rollouts and teach the action grammar. The main reasoning behavior is learned through RL.
Starting from the cold-start policy, we train with the action-local rewards defined above. In early RL, we use easier questions and data with shorter context to stabilize exploration. 
% In the full stage, the policy is trained with all reward components.
We cap the number of add, drop, and link actions to prevent degenerate loops and to bound verifier cost. Additional methodology illustration, including details of the curriculum training strategy, the choice of the maximum number of allowed actions, and computational analysis are presented in Appendix~\ref{app:method}.
\section{Experiments}
\label{sec:experiment}
\subsection{Setup}
\paragraph{Training Data.}
For RL training, we train on 9K long-context question--answer pairs with contexts from 8K to 64K tokens. The set contains 3K examples from LongRLVR~\citep{chen2026longrlvr} and 6K multi-hop examples from HotpotQA~\citep{yang2018hotpotqa}, 2WikiMultiHopQA~\citep{ho2020constructing}, and MuSiQue~\citep{trivedi2022musique}. For LongRLVR data, we select examples that require multiple evidence chunks to answer and filter candidates according to context length, number of necessary evidence units, absence of single-evidence shortcuts, and question difficulty. 
% For the multi-hop datasets, we construct long contexts by combining gold supporting evidence with distractors. In addition to random distractors and dataset-provided distractors, we use Qwen3-235B-A22B~\cite{yang2025qwen3} and Gemini 3 to generate hard distractors that are lexically or semantically related to the question but do not support the gold answer. In particular, we ask the teacher models to construct partial-chain distractors, which share entities, relations, or answer types with the gold reasoning chain but lead to an incorrect conclusion. These distractors are designed to elicit dropping behavior during RL training. We discard distractors that reveal the gold answer, duplicate gold evidence, or make the answer ambiguous, and then apply the same filtering procedure used for the original examples. 
For the multi-hop data, we combine gold evidence with random, dataset-provided, and model-generated hard distractors.
% For cold-start SFT, we construct 2K examples from the same source datasets. We prompt Qwen3-235B-A22B to generate trajectories following the editable evidence-memory format with \texttt{add}, \texttt{link}, \texttt{drop}, and \texttt{answer} actions. 
For cold-start SFT, we construct 2K trajectories in the editable evidence-memory format from the same source datasets.
Details of data sampling, distractor construction, filtering criteria, and data statistics are provided in Appendix~\ref{app:data}.

\paragraph{Implementation Setup.}
We conduct experiments on three policy models from different model families and scales: Llama-3.1-8B-Instruct~\citep{dubey2024llama3herdmodels}, Qwen2.5-14B-Instruct~\citep{Yang2024Qwen25TR}, and Qwen3-30B-A3B-Instruct-2507~\citep{yang2025qwen3}. For cold-start SFT, we use a learning rate of \(2\times 10^{-5}\), a batch size of 16, and 30 warmup steps. For RL training, we use GRPO with a rollout group size of 8 and a global prompt batch size of 16. We train for one epoch with a learning rate of \(1\times 10^{-6}\), cosine decay, and 10 warmup steps. The maximum response length is 4096 tokens, and rollouts are sampled with \texttt{temperature=0.8} and \texttt{top\_p=0.95}. We use Qwen3-4B-Instruct-2507~\citep{yang2025qwen3} as the frozen verifier. Unless otherwise specified, we set the reward weights of add, drop, link, and answer rewards to \(0.5\), \(0.2\), \(0.3\), and \(1.0\), respectively, and set \(\alpha=0.6\) in the add reward.

\paragraph{Baselines and Benchmarks.}
We compare \textsc{Maven} with several controlled baselines trained under the same data and optimization setup: the original base model, the cold-start SFT model, outcome-only RL using final answer correctness as reward, outcome plus evidence identification reward, and outcome plus evidence identification reward with an exploration-oriented prompt. The evidence identification reward is computed as the F1 score between the model's final kept evidence set and the gold evidence set. The prompted exploration baseline uses the same action format as \textsc{Maven}, but does not reward add, drop, or link actions separately. The rollout prompt for outcome-only baseline instructs the model to output the final answer wrapped by \texttt{<answer>} and \texttt{</answer>}. In the rollout prompt for outcome+evidence identification baseline, we instruct the model to explicitly indicate the ID of selected chunks. We additionally report results of stronger open-weight models, including Llama-3.1-70B~\cite{dubey2024llama3herdmodels}, Qwen3-32B (Thinking)~\citep{yang2025qwen3}, and QwenLong-L1-32B~\citep{wan2025qwenlong}, as reference points. We use YaRN~\citep{peng2024yarn} to extend the context length of Qwen models to 128K when needed. We evaluate all models on three long-context benchmarks: \textbf{LongBench v2}~\citep{bai2025longbench}, \textbf{LongReason}~\citep{ling2025longreason}, and selected subsets from \textbf{RULER}~\citep{hsieh2024ruler}. All evaluations follow the official inference configuration of each benchmark. 
% In addition, we use a held-out diagnostic set of 150 constructed examples with gold evidence annotations to evaluate evidence-process behavior.
% \paragraph{Diagnostic Evaluation.}
In addition to benchmark accuracy, we evaluate evidence-process behavior on a held-out diagnostic set of 150 constructed examples from multi-hop data with gold evidence annotations. 
We report evidence sufficiency and distractor retention. Evidence sufficiency measures the fraction of gold evidence chunks that are covered in the final kept evidence. Distractor retention measures the percentage of final kept evidence items that correspond to distractor chunks. For diagnostic evaluation, all models are prompted to output evidence identifiers so that evidence behavior can be measured consistently.

We provide additional implementation and evaluation details in Appendix~\ref{app:implementation}.

\subsection{Main Results}
\definecolor{avgblue}{HTML}{EAF3FF}
\definecolor{bestgreen}{HTML}{E6F4EA}
\definecolor{cellorange}{HTML}{FFF2CC}

\begin{table*}[t]
\centering
\small
\caption{Main results on LongBench v2, LongReason, and RULER. Bold numbers indicate the best method for each trained model. Green cells indicate the best result among all listed models.}
\label{tab:main_results}
\resizebox{0.95\textwidth}{!}{%
\begin{tabular}{l
cccc
cccc
ccc
}
\toprule
\multirow{2}{*}{\textbf{Model}} &  \multicolumn{4}{c}{\textbf{LongBench v2}} & \multicolumn{4}{c}{\textbf{LongReason}} &\multicolumn{3}{c}{\textbf{RULER}}  \\
\cmidrule(lr){2-5} \cmidrule(lr){6-9} \cmidrule(lr){10-12}
& {\textbf{Short}} & {\textbf{Medium}} & {\textbf{Long}} & {\textbf{Overall}} & {\textbf{32K}} & {\textbf{64K}} & {\textbf{128K}} & {\textbf{AVG}} &    {\textbf{64K}} & {\textbf{128K}} & {\textbf{Avg.}} \\
\midrule
LLaMA-3.1-70B &42.8&38.0&31.2&38.3&61.2&63.3&48.3&57.6&93.2&69.8&81.5 \\
Qwen3-32B (Thinking) &\cellcolor{bestgreen}{56.7}&44.0&45.1&48.7&\cellcolor{bestgreen}86.6&84.4&79.3&83.5&92.1&84.4& 88.2\\
QwenLong-L1-32B &52.8&36.2&32.7&41.4&84.1&83.6&75.1&80.9&81.7&74.3& 78.0\\
\midrule
\midrule
LLaMA-3.1-8B &33.3&30.7&22.5&29.9&51.4&49.9&46.5&49.3&85.1&77.2& 81.1\\
\quad + SFT &34.5&30.4&24.1&30.5&51.0&49.2&47.1&49.1&85.6&77.4& 81.5 \\
\quad + Outcome &36.7&32.0&24.4&32.0&51.8&50.5&46.2&49.5&85.8&77.9& 81.9 \\
\quad + Outcome+Evidence ID &38.1&32.6&25.3&33.0&52.1&50.9&48.7&50.6&86.9&78.3& 82.6 \\
\quad + Outcome+Evidence ID (Prompted exploration) &37.6&32.9&25.9&33.1&52.4&51.3&48.9&50.8&86.6&78.5& 82.5 \\
\quad + \textbf{\textsc{MAVEN}} &\textbf{39.8}&\textbf{36.2}&\textbf{32.1}&\textbf{36.6}&\textbf{55.9}&\textbf{55.2}&\textbf{55.8}&\textbf{55.6}&\textbf{88.4}&\textbf{80.1}& \textbf{84.3} \\
\midrule
Qwen2.5-14B &47.6&33.9&30.2&38.0&68.1&66.2&62.3&65.5&83.7&75.5& 79.6\\
\quad + SFT &46.8&34.7&31.2&38.3&67.6&66.8&61.5&65.3&83.9&75.2& 79.6 \\
\quad + Outcome &48.1&34.9&31.2&38.8&69.5&67.4&62.0&66.3&84.1&75.7& 79.9 \\
\quad + Outcome+Evidence ID &49.3&36.1&33.0&40.1&70.3&67.8&64.7&67.6&86.7&76.9& 81.8 \\
\quad + Outcome+Evidence ID (Prompted exploration) &49.3&36.5&33.0&40.3&70.2&67.6&64.5&67.4&86.5&77.1& 81.8 \\
\quad + \textbf{\textsc{MAVEN}} &\textbf{51.5}&\textbf{40.2}&\textbf{37.0}&\textbf{43.5}&\textbf{73.0}&\textbf{71.7}&\textbf{70.2}&\textbf{71.6}&\textbf{90.0}&\textbf{81.6}&\textbf{85.8} \\
\midrule
Qwen3-30B-A3B &50.7&39.4&40.7&43.7&84.8&82.9&77.1&81.6&88.2&82.6& 85.4 \\
\quad + SFT &49.3&38.1&38.9&42.2&83.8&82.6&76.3&80.9&88.5&82.7& 85.6 \\
\quad + Outcome &49.8&39.7&40.7&43.5&84.8&81.7&77.2&80.9&87.2&82.3&84.8 \\
\quad + Outcome+Evidence ID &48.9&41.1&44.4&44.6&85.3&82.6&79.3&82.4&89.6&84.4&87.0 \\
\quad + Outcome+Evidence ID (Prompted exploration) &49.3&41.6&43.8&44.8&85.6&82.5&79.6&82.6&89.8&84.5&87.2\\
\quad + \textbf{\textsc{MAVEN}} &\textbf{53.9}&\cellcolor{bestgreen}\textbf{45.9}&\cellcolor{bestgreen}\textbf{46.3}&\cellcolor{bestgreen}\textbf{48.8}&\cellcolor{bestgreen}\textbf{86.6}&\cellcolor{bestgreen}\textbf{85.1}&\cellcolor{bestgreen}\textbf{81.7}&\cellcolor{bestgreen}\textbf{84.5}&\cellcolor{bestgreen}\textbf{93.4}&\cellcolor{bestgreen}\textbf{88.8}&\cellcolor{bestgreen}\textbf{91.1} \\
\bottomrule
\end{tabular}
}
\end{table*}

Table~\ref{tab:main_results} reports the main results on LongBench v2, LongReason, and RULER. \textsc{Maven} consistently outperforms all controlled baselines across model families, model sizes, and benchmarks. On LongBench v2 overall, \textsc{Maven} improves over the strongest baseline by \(+3.5\), \(+3.2\), and \(+4.0\) points on Llama-3.1-8B, Qwen2.5-14B, and Qwen3-30B-A3B, respectively. Similar gains are observed on LongReason and RULER, showing that the proposed evidence-memory rewards transfer beyond the training data distribution.

The improvements are especially clear in long-context settings. On the LongBench v2 Long split, \textsc{Maven} improves Llama-3.1-8B from \(25.9\) to \(32.1\), Qwen2.5-14B from \(33.0\) to \(37.0\), and Qwen3-30B-A3B from \(43.8\) to \(46.3\), compared with the strongest baseline for each model. On LongReason, \textsc{Maven} also gives substantial gains at 128K context length.
These results suggest that modeling the evolving evidence state is particularly beneficial when the answer depends on information distributed across long contexts.

Compared with outcome-only RL, the Outcome+Evidence ID baseline consistently improves performance, confirming that dense evidence supervision is useful for long-context RL. However, it remains notably below \textsc{Maven}. This gap indicates that rewarding final evidence overlap alone is insufficient: the model also needs feedback on how evidence is added, revised, and synthesized during the reasoning process. The prompted exploration baseline gives only marginal additional gains over Outcome+Evidence ID, suggesting that simply instructing the model to explore is less effective than explicitly rewarding useful evidence-state transitions.

Among all listed models, Qwen3-30B-A3B trained with \textsc{Maven} achieves the best or tied-best result on most reported columns, including the overall scores of all three benchmarks. Notably, on RULER, Llama-3.1-8B trained with \textsc{Maven} surpasses the Llama-3.1-70B reference model on the evaluated subsets. Meanwhile, Qwen3-30B-A3B already performs strongly before training, but \textsc{Maven} further improves its RULER average from \(85.4\) to \(91.1\), showing that the proposed method remains beneficial even for strong long-context models.

\subsection{Training Dynamics}
\label{sec:training_dynamics}
\begin{figure*}[t]
    \centering
    \includegraphics[width=1.0\textwidth]{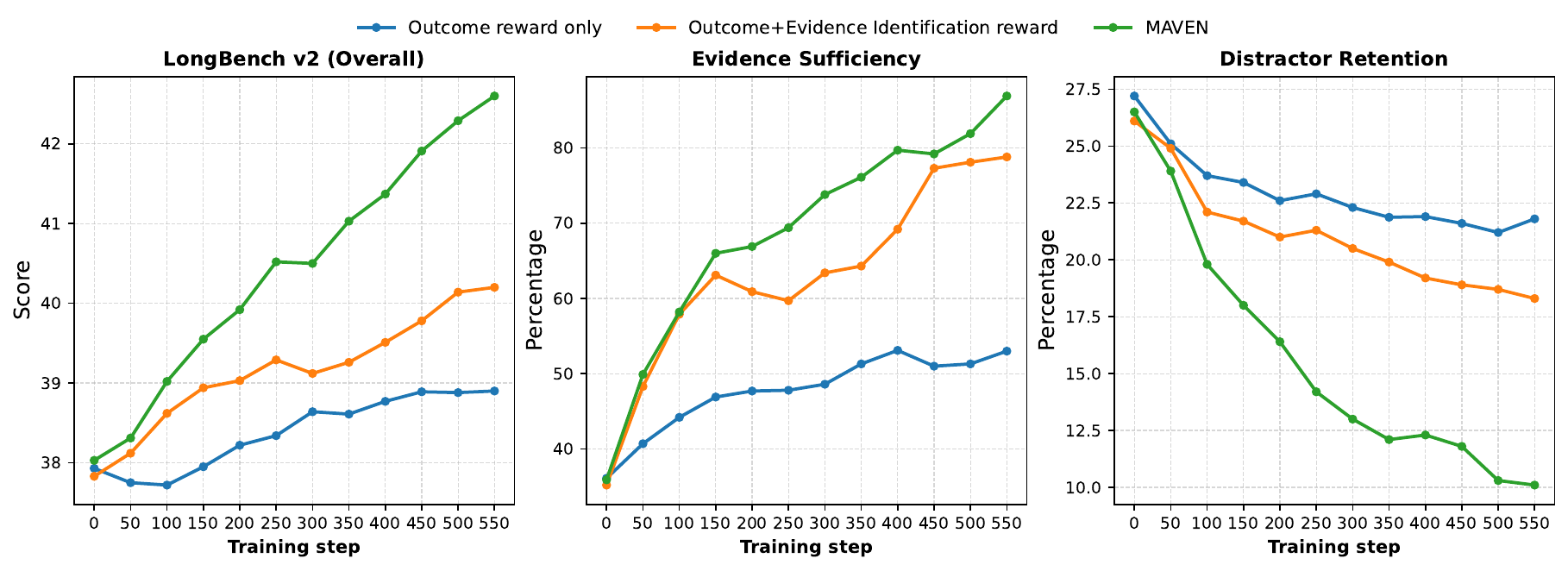}
    \caption{Training dynamics of outcome-only RL, Outcome+Evidence ID, and \textsc{Maven}. We evaluate every 50 RL steps on a fixed subset of LongBench v2 examples and on the diagnostic set. 
    }
    \label{fig:reward_comparison}
\end{figure*}

Figure~\ref{fig:reward_comparison} compares training dynamics of Qwen2.5-14B-Instruct across outcome-only RL, Outcome+Evidence ID, and \textsc{Maven}. We evaluate every 50 steps on a fixed subset of 120 LongBench v2 examples and on the diagnostic set. The monitoring subset is used only for analysis, not for checkpoint selection. 

Outcome-only RL improves slowly and saturates early. Adding evidence identification reward leads to stronger improvement. \textsc{Maven} improves more steadily throughout training and shows a higher performance ceiling.

The diagnostic curves explain this gap. Outcome+Evidence ID substantially improves evidence sufficiency, showing that evidence identification reward can guide the model toward selecting relevant chunks. However, its distractor retention remains relatively high. In contrast, \textsc{Maven} improves evidence sufficiency to over 85 and significantly reduces distractor retention. This suggests that \textsc{Maven} not only teaches the model to find useful evidence, but also trains it to revise the evidence memory and discard misleading chunks.

\subsection{Impact of Action Rewards}
\label{sec:experiment_action_rewards}
\begin{figure*}[t]
    \centering
    \includegraphics[width=0.95\textwidth]{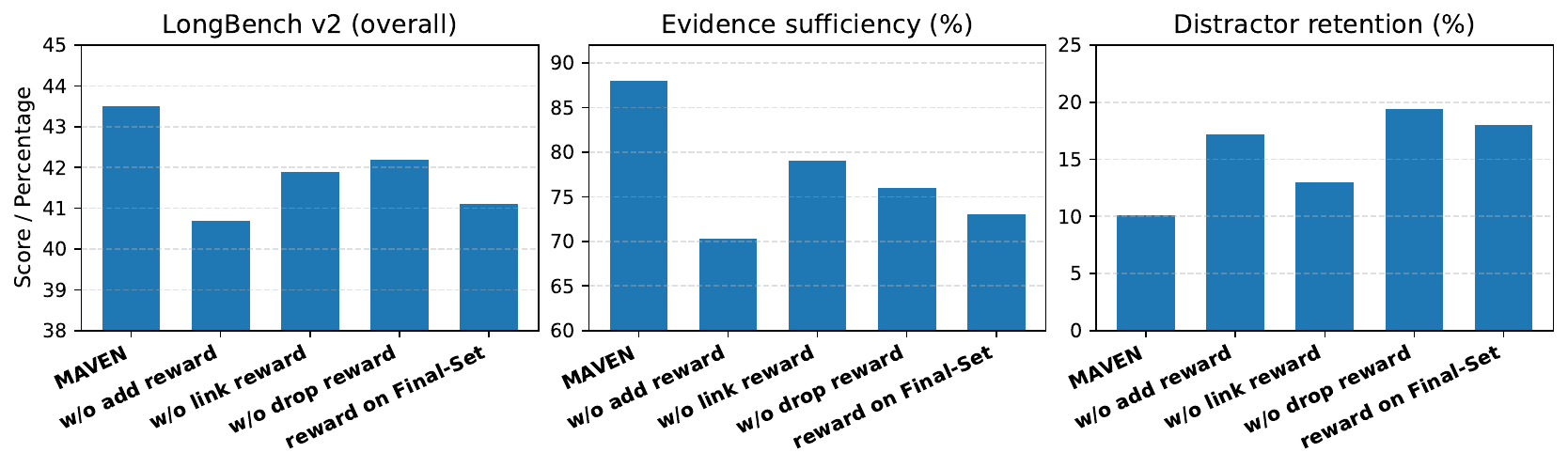}
    \caption{Impact of action rewards. We measure LongBench V2 overall score, evidence sufficiency, and distractor retention.}
    \label{fig:ablation}
\end{figure*}

Figure~\ref{fig:ablation} studies the impact of different action rewards in our method on Qwen2.5-14B-Instruct. Removing the add reward causes the largest drop in LongBench v2 overall score. 
% This confirms that online Conditional Answer Information Gain and hindsight evidence credit are central to learning effective evidence construction.
The link and drop rewards also contribute complementary benefits. Removing the link reward lowers evidence sufficiency, suggesting that explicit synthesis rewards help the model combine complementary evidence pieces. Removing the drop reward leads to the highest distractor retention, which confirms that drop supervision is important for evidence revision and distractor removal. We also compare against a final-evidence-only variant, where the add, link and drop reward is computed only from the final kept evidence set. This variant achieves scores that are substantially below full \textsc{Maven}. This shows that action-local process rewards are important for learning how to build and revise the evidence memory.

Interestingly, the effects of action rewards are not isolated to their corresponding behaviors. For example, removing the drop reward also reduces evidence sufficiency, and removing the add reward increases distractor retention. This supports our central claim that long-context reasoning is a dynamic process involving interactions among different actions. Additional action-level diagnostics, including add precision, drop precision, and link precision, are provided in Appendix~\ref{app:ablation}.

\subsection{Further Analysis}
\label{sec:further_analysis}
\paragraph{Choice of Verifier.} Table~\ref{tab:verifier_results} studies the effect of verifier choice. We compare the default Qwen3-4B verifier with a Qwen2.5-7B-Instruct verifier on Llama-3.1-8B and Qwen2.5-14B. The results are nearly unchanged. This suggests that \textsc{Maven} is not highly sensitive to the specific frozen verifier used for answer-NLL scoring. We therefore use the smaller Qwen3-4B verifier in the main experiments for better efficiency.

\paragraph{Contrastive Answer Scoring.}
The main method computes evidence-state value using the teacher-forced probability of the gold answer over the full vocabulary. One concern is that the verifier's prior knowledge may introduce noise into this value estimate. To examine this, we construct a contrastive answer set for each training example, containing the gold answer, several similar but incorrect answers, and an abstention option. We then compute the value using the normalized probability of the gold answer within this contrastive set. Results of Qwen2.5-14B are shown in Table~\ref{tab:token_prob}. Contrastive scoring yields a small improvement, but the gain is limited relative to its additional construction cost. Therefore, we use full-vocabulary answer-token NLL in the main method as an efficient approximation.
 \begin{table}
    \centering
    \small
    \caption{Comparison with scoring using contrastive answers on LongBench v2.} 
    \label{tab:token_prob}
    \resizebox{0.95\linewidth}{!}{
    \begin{tabular}{lcccc} 
        \toprule
        & Short & Medium & Long & Overall  \\
        \midrule
    whole vocabulary  & 51.5 & 40.2 & 37.0 & 43.5\\            
      contrastive answer set & 51.6& 41.1 &  37.6 & 44.1 \\
  
        \bottomrule
    \end{tabular}
    }
\end{table}

\paragraph{Reward Hyperparameters.}
We study the sensitivity of \textsc{Maven} to reward weights and to \(\alpha\), which balances online CAIG and hindsight credit in the add reward. Overall, the best performance is achieved around our default setting.
% :
% \(\lambda_{\mathrm{ans}}=1.0\), \(\lambda_{\mathrm{add}}=0.5\), \(\lambda_{\mathrm{drop}}=0.2\), \(\lambda_{\mathrm{link}}=0.3\), and \(\alpha=0.6\).
The results show a trade-off between answer supervision and process supervision: underweighting process rewards weakens evidence construction, while overweighting them can distract the model from final answer correctness. Increasing the add weight generally improves evidence sufficiency, and increasing the drop weight reduces distractor retention, but overly aggressive evidence addition or removal can hurt overall performance. Detailed results are provided in Appendix~\ref{app:hyperparameter}.

\paragraph{Performance on General Short-Context Reasoning.}
To examine whether long-context RL training harms general short-context reasoning ability, we evaluate the base and \textsc{Maven}-trained models on MMLU-Pro~\citep{wang2024mmlu}. As shown in Table~\ref{tab:mmlu}, \textsc{Maven} does not lead to clear degradation, suggesting that the proposed method preserves the models' general reasoning capability.

\section{Related Work}

\paragraph{Long-context grounding and reasoning.} 
Prior work has addressed long contexts by extending the usable context window through architectural or positional methods~\citep{chen2023extending,su2024roformer,ding2024longrope,team2025kimi}, reducing the effective input length with retrieval~\citep{lewis2020retrieval,jiang2024longrag,zhao2024longrag}, or decomposing long inputs with agentic workflows~\citep{zhang2024chain}. These approaches improve access to information, but they do not necessarily teach the model to reason over long inputs. A closer line of work trains models directly for long-context behavior using supervised finetuning~\citep{bai2024longalign} or reinforcement learning~\citep{wan2025qwenlong,wang2025loongrl,chen2026longrlvr}. Existing long-context RL methods commonly optimize final-answer rewards~\citep{wan2025qwenlong,zhang2025longreward,wang2025loongrl} or add context-aware rewards that score selected chunks~\citep{chen2026longrlvr,guan2026evidence,ping2026longr}. While context-aware rewards provide denser feedback than outcome-only supervision, they largely remain extraction-centric. In contrast, \textsc{Maven} treats long-context reasoning as evidence-state navigation: it rewards how different actions change an answer-conditioned evidence memory. Our method is therefore complementary to retrieval, agentic workflows, and architecture-level context extension, while targeting a different bottleneck: learning to build and revise the evidence state used for reasoning.

\paragraph{Reward design and process credit assignment.}
RL with verifiable rewards has improved LLM reasoning~\citep{jaech2024openai,guo2025deepseek}, but final-answer rewards provide only trajectory-level feedback. Process supervision assigns feedback to intermediate steps~\citep{lightman2024let,zhang2025lessons,khalifa2025process}; however, long-context reasoning requires process feedback over evidence editing, not only free-form reasoning traces. \textsc{Maven} assigns action-local advantages to add, link, drop, and answer spans using an answer-conditioned evidence-state value, distinguishing it from static evidence rewards and learned black-box process reward models.

\section{Conclusion}
We presented \textsc{Maven}, a reinforcement learning framework that treats long-context reasoning as stateful evidence-memory navigation. By using an answer-conditioned evidence-state value and add, link, and drop rewards, \textsc{Maven} provides process supervision for building, revising, and synthesizing evidence rather than only rewarding final answers or static evidence overlap. Across Llama and Qwen models on three benchmarks, \textsc{Maven} consistently improves over RL baselines. Analyses further show higher evidence sufficiency, lower distractor retention, and complementary benefits from the three action rewards. These results suggest that effective long-context RL should optimize evidence-state transitions, not merely evidence extraction.

\section*{Limitations}
\textsc{Maven} relies on gold answers during training to compute verifier-based evidence-state values, so it is most directly applicable to tasks with verifiable answers. The editable evidence interface also assumes that evidence can be represented as bounded text spans. Extensions to multimodal evidence, or open-ended generation may require additional action designs and evaluation protocols. Finally, the diagnostic evidence annotations are constructed for controlled analysis, so broader human evaluation of evidence quality and faithfulness remains important future work.

\bibliography{main}

@article{guo2025deepseek,
  title={Deepseek-r1: Incentivizing reasoning capability in llms via reinforcement learning},
  author={Guo, Daya and Yang, Dejian and Zhang, Haowei and Song, Junxiao and Wang, Peiyi and Zhu, Qihao and Xu, Runxin and Zhang, Ruoyu and Ma, Shirong and Bi, Xiao and others},
  journal={arXiv preprint arXiv:2501.12948},
  year={2025}
}

@article{jaech2024openai,
  title={Openai o1 system card},
  author={Jaech, Aaron and Kalai, Adam and Lerer, Adam and Richardson, Adam and El-Kishky, Ahmed and Low, Aiden and Helyar, Alec and Madry, Aleksander and Beutel, Alex and Carney, Alex and others},
  journal={arXiv preprint arXiv:2412.16720},
  year={2024}
}

@article{team2025kimiagent,
  title={Kimi k2: Open agentic intelligence},
  author={Team, Kimi and Bai, Yifan and Bao, Yiping and Charles, Y and Chen, Cheng and Chen, Guanduo and Chen, Haiting and Chen, Huarong and Chen, Jiahao and Chen, Ningxin and others},
  journal={arXiv preprint arXiv:2507.20534},
  year={2025}
}

@article{liu2024lost,
  title={Lost in the middle: How language models use long contexts},
  author={Liu, Nelson F and Lin, Kevin and Hewitt, John and Paranjape, Ashwin and Bevilacqua, Michele and Petroni, Fabio and Liang, Percy},
  journal={Transactions of the association for computational linguistics},
  volume={12},
  pages={157--173},
  year={2024}
}

@article{wang2025loongrl,
  title={Loongrl: Reinforcement learning for advanced reasoning over long contexts},
  author={Wang, Siyuan and Zhang, Gaokai and Zhang, Li Lyna and Shang, Ning and Yang, Fan and Chen, Dongyao and Yang, Mao},
  journal={arXiv preprint arXiv:2510.19363},
  year={2025}
}

@inproceedings{
  chen2026longrlvr,
  title={Long{RLVR}: Long-Context Reinforcement Learning Requires Verifiable Context Rewards},
  author={Guanzheng Chen and Michael Qizhe Shieh and Lidong Bing},
  booktitle={The Fourteenth International Conference on Learning Representations},
  year={2026},
  url={https://openreview.net/forum?id=omVhYvyTPJ}
}

@article{shao2024deepseekmath,
  title={Deepseekmath: Pushing the limits of mathematical reasoning in open language models},
  author={Shao, Zhihong and Wang, Peiyi and Zhu, Qihao and Xu, Runxin and Song, Junxiao and Bi, Xiao and Zhang, Haowei and Zhang, Mingchuan and Li, YK and Wu, Yang and others},
  journal={arXiv preprint arXiv:2402.03300},
  year={2024}
}

@article{team2025kimi,
  title={Kimi linear: An expressive, efficient attention architecture},
  author={Team, Kimi and Zhang, Yu and Lin, Zongyu and Yao, Xingcheng and Hu, Jiaxi and Meng, Fanqing and Liu, Chengyin and Men, Xin and Yang, Songlin and Li, Zhiyuan and others},
  journal={arXiv preprint arXiv:2510.26692},
  year={2025}
}

@article{wan2025qwenlong,
  title={Qwenlong-l1: Towards long-context large reasoning models with reinforcement learning},
  author={Wan, Fanqi and Shen, Weizhou and Liao, Shengyi and Shi, Yingcheng and Li, Chenliang and Yang, Ziyi and Zhang, Ji and Huang, Fei and Zhou, Jingren and Yan, Ming},
  journal={arXiv preprint arXiv:2505.17667},
  year={2025}
}

@article{guan2026evidence,
  title={Evidence-Augmented Policy Optimization with Reward Co-Evolution for Long-Context Reasoning},
  author={Guan, Xin and Li, Zijian and Huang, Shen and Xie, Pengjun and Zhou, Jingren and Cao, Jiuxin},
  journal={arXiv preprint arXiv:2601.10306},
  year={2026}
}

@article{ping2026longr,
  title={LongR: Unleashing Long-Context Reasoning via Reinforcement Learning with Dense Utility Rewards},
  author={Ping, Bowen and Chen, Zijun and Yu, Yiyao and Hui, Tingfeng and Yan, Junchi and Chang, Baobao},
  journal={arXiv preprint arXiv:2602.05758},
  year={2026}
}

@inproceedings{yang2018hotpotqa,
  title={HotpotQA: A dataset for diverse, explainable multi-hop question answering},
  author={Yang, Zhilin and Qi, Peng and Zhang, Saizheng and Bengio, Yoshua and Cohen, William and Salakhutdinov, Ruslan and Manning, Christopher D},
  booktitle={Proceedings of the 2018 conference on empirical methods in natural language processing},
  pages={2369--2380},
  year={2018}
}

@article{trivedi2022musique,
  title={MuSiQue: Multihop Questions via Single-hop Question Composition},
  author={Trivedi, Harsh and Balasubramanian, Niranjan and Khot, Tushar and Sabharwal, Ashish},
  journal={Transactions of the Association for Computational Linguistics},
  volume={10},
  pages={539--554},
  year={2022},
  publisher={MIT Press One Broadway, 12th Floor, Cambridge, Massachusetts 02142, USA~…}
}

@inproceedings{ho2020constructing,
  title={Constructing a multi-hop qa dataset for comprehensive evaluation of reasoning steps},
  author={Ho, Xanh and Nguyen, Anh-Khoa Duong and Sugawara, Saku and Aizawa, Akiko},
  booktitle={Proceedings of the 28th International Conference on Computational Linguistics},
  pages={6609--6625},
  year={2020}
}

@article{yang2025qwen3,
  title={Qwen3 technical report},
  author={Yang, An and Li, Anfeng and Yang, Baosong and Zhang, Beichen and Hui, Binyuan and Zheng, Bo and Yu, Bowen and Gao, Chang and Huang, Chengen and Lv, Chenxu and others},
  journal={arXiv preprint arXiv:2505.09388},
  year={2025}
}

@inproceedings{bai2025longbench,
  title={Longbench v2: Towards deeper understanding and reasoning on realistic long-context multitasks},
  author={Bai, Yushi and Tu, Shangqing and Zhang, Jiajie and Peng, Hao and Wang, Xiaozhi and Lv, Xin and Cao, Shulin and Xu, Jiazheng and Hou, Lei and Dong, Yuxiao and others},
  booktitle={Proceedings of the 63rd Annual Meeting of the Association for Computational Linguistics (Volume 1: Long Papers)},
  pages={3639--3664},
  year={2025}
}

@article{ling2025longreason,
  title={Longreason: A synthetic long-context reasoning benchmark via context expansion},
  author={Ling, Zhan and Liu, Kang and Yan, Kai and Yang, Yifan and Lin, Weijian and Fan, Ting-Han and Shen, Lingfeng and Du, Zhengyin and Chen, Jiecao},
  journal={arXiv preprint arXiv:2501.15089},
  year={2025}
}

@article{hsieh2024ruler,
  title={RULER: What's the real context size of your long-context language models?},
  author={Hsieh, Cheng-Ping and Sun, Simeng and Kriman, Samuel and Acharya, Shantanu and Rekesh, Dima and Jia, Fei and Zhang, Yang and Ginsburg, Boris},
  journal={arXiv preprint arXiv:2404.06654},
  year={2024}
}

@inproceedings{peng2024yarn,
  title={Yarn: Efficient context window extension of large language models},
  author={Peng, Bowen and Quesnelle, Jeffrey and Fan, Honglu and Shippole, Enrico},
  booktitle={International Conference on Learning Representations},
  volume={2024},
  pages={31932--31951},
  year={2024}
}

@article{jin2025search,
  title={Search-r1: Training llms to reason and leverage search engines with reinforcement learning},
  author={Jin, Bowen and Zeng, Hansi and Yue, Zhenrui and Yoon, Jinsung and Arik, Sercan and Wang, Dong and Zamani, Hamed and Han, Jiawei},
  journal={arXiv preprint arXiv:2503.09516},
  year={2025}
}

@article{su2024roformer,
  title={Roformer: Enhanced transformer with rotary position embedding},
  author={Su, Jianlin and Ahmed, Murtadha and Lu, Yu and Pan, Shengfeng and Bo, Wen and Liu, Yunfeng},
  journal={Neurocomputing},
  volume={568},
  pages={127063},
  year={2024},
  publisher={Elsevier}
}

@article{chen2023extending,
  title={Extending context window of large language models via positional interpolation},
  author={Chen, Shouyuan and Wong, Sherman and Chen, Liangjian and Tian, Yuandong},
  journal={arXiv preprint arXiv:2306.15595},
  year={2023}
}

@article{ding2024longrope,
  title={Longrope: Extending llm context window beyond 2 million tokens},
  author={Ding, Yiran and Zhang, Li Lyna and Zhang, Chengruidong and Xu, Yuanyuan and Shang, Ning and Xu, Jiahang and Yang, Fan and Yang, Mao},
  journal={arXiv preprint arXiv:2402.13753},
  year={2024}
}

@article{lewis2020retrieval,
  title={Retrieval-augmented generation for knowledge-intensive nlp tasks},
  author={Lewis, Patrick and Perez, Ethan and Piktus, Aleksandra and Petroni, Fabio and Karpukhin, Vladimir and Goyal, Naman and K{\"u}ttler, Heinrich and Lewis, Mike and Yih, Wen-tau and Rockt{\"a}schel, Tim and others},
  journal={Advances in neural information processing systems},
  volume={33},
  pages={9459--9474},
  year={2020}
}

@article{jiang2024longrag,
  title={Longrag: Enhancing retrieval-augmented generation with long-context llms},
  author={Jiang, Ziyan and Ma, Xueguang and Chen, Wenhu},
  journal={arXiv preprint arXiv:2406.15319},
  year={2024}
}

@inproceedings{zhao2024longrag,
  title={Longrag: A dual-perspective retrieval-augmented generation paradigm for long-context question answering},
  author={Zhao, Qingfei and Wang, Ruobing and Cen, Yukuo and Zha, Daren and Tan, Shicheng and Dong, Yuxiao and Tang, Jie},
  booktitle={Proceedings of the 2024 Conference on Empirical Methods in Natural Language Processing},
  pages={22600--22632},
  year={2024}
}

@article{zhang2024chain,
  title={Chain of agents: Large language models collaborating on long-context tasks},
  author={Zhang, Yusen and Sun, Ruoxi and Chen, Yanfei and Pfister, Tomas and Zhang, Rui and Ar{\i}k, Sercan {\"O}},
  journal={Advances in Neural Information Processing Systems},
  volume={37},
  pages={132208--132237},
  year={2024}
}

@inproceedings{bai2024longalign,
  title={Longalign: A recipe for long context alignment of large language models},
  author={Bai, Yushi and Lv, Xin and Zhang, Jiajie and He, Yuze and Qi, Ji and Hou, Lei and Tang, Jie and Dong, Yuxiao and Li, Juanzi},
  booktitle={Findings of the Association for Computational Linguistics: EMNLP 2024},
  pages={1376--1395},
  year={2024}
}

@inproceedings{zhang2025longreward,
  title={Longreward: Improving long-context large language models with ai feedback},
  author={Zhang, Jiajie and Hou, Zhongni and Lv, Xin and Cao, Shulin and Hou, Zhenyu and Niu, Yilin and Hou, Lei and Dong, Yuxiao and Feng, Ling and Li, Juanzi},
  booktitle={Proceedings of the 63rd Annual Meeting of the Association for Computational Linguistics (Volume 1: Long Papers)},
  pages={3718--3739},
  year={2025}
}

@article{khalifa2025process,
  title={Process reward models that think},
  author={Khalifa, Muhammad and Agarwal, Rishabh and Logeswaran, Lajanugen and Kim, Jaekyeom and Peng, Hao and Lee, Moontae and Lee, Honglak and Wang, Lu},
  journal={arXiv preprint arXiv:2504.16828},
  year={2025}
}

@inproceedings{zhang2025lessons,
  title={The lessons of developing process reward models in mathematical reasoning},
  author={Zhang, Zhenru and Zheng, Chujie and Wu, Yangzhen and Zhang, Beichen and Lin, Runji and Yu, Bowen and Liu, Dayiheng and Zhou, Jingren and Lin, Junyang},
  booktitle={Findings of the Association for Computational Linguistics: ACL 2025},
  pages={10495--10516},
  year={2025}
}

@inproceedings{lightman2024let,
  title={Let's verify step by step},
  author={Lightman, Hunter and Kosaraju, Vineet and Burda, Yuri and Edwards, Harrison and Baker, Bowen and Lee, Teddy and Leike, Jan and Schulman, John and Sutskever, Ilya and Cobbe, Karl},
  booktitle={International Conference on Learning Representations},
  volume={2024},
  pages={39578--39601},
  year={2024}
}

@article{wang2024mmlu,
  title={Mmlu-pro: A more robust and challenging multi-task language understanding benchmark},
  author={Wang, Yubo and Ma, Xueguang and Zhang, Ge and Ni, Yuansheng and Chandra, Abhranil and Guo, Shiguang and Ren, Weiming and Arulraj, Aaran and He, Xuan and Jiang, Ziyan and others},
  journal={Advances in Neural Information Processing Systems},
  volume={37},
  pages={95266--95290},
  year={2024}
}

@misc{dubey2024llama3herdmodels,
      title={The Llama 3 Herd of Models}, 
      author={Abhimanyu Dubey and Abhinav Jauhri and Abhinav Pandey and Abhishek Kadian and Alon Al-Dahle and Allan Letman and Andy Mathur and Angie Schelten and Anthony Yang and Armin Fan and Aruna Goyal and Bram Roziere and Branislav Biron and Charbel Bitar and Chantat Ng and Chen Xia and Cheng Tan and Christian Keller and Christian Touret and Conceicao de Clercq and Dmytro Okhonko and Dorian Esiobu and ...},
      year={2024},
      eprint={2407.21783},
      archivePrefix={arXiv},
      primaryClass={cs.CL}
}

@article{Yang2024Qwen25TR,
  title={Qwen2.5 Technical Report},
  author={Qwen An Yang and Baosong Yang and Beichen Zhang and Binyuan Hui and Bo Zheng and Bowen Yu and Chengyuan Li and Dayiheng Liu and Fei Huang and Guanting Dong and Haoran Wei and Huan Lin and Jian Yang and Jianhong Tu and Jianwei Zhang and Jianxin Yang and Jiaxin Yang and Jingren Zhou and Junyang Lin and Kai Dang and Keming Lu and Keqin Bao and Kexin Yang and Le Yu and Mei Li and Mingfeng Xue and Pei Zhang and Qin Zhu and Rui Men and Runji Lin and Tianhao Li and Tingyu Xia and Xingzhang Ren and Xuancheng Ren and Yang Fan and Yang Su and Yi-Chao Zhang and Yunyang Wan and Yuqi Liu and Zeyu Cui and Zhenru Zhang and Zihan Qiu and Shanghaoran Quan and Zekun Wang},
  journal={ArXiv},
  year={2024},
  volume={abs/2412.15115},
  url={https://api.semanticscholar.org/CorpusID:274859421}
}

\newpage

\appendix

\section{Appendix}
\subsection{Additional Methodology Details}
\label{app:method}
\subsubsection{Curriculum RL Training Strategy}
We apply a simple curriculum strategy during the early stage of RL training. In the first 20\% training updates under our default batch size, we exclude examples whose context length exceeds 16K tokens and use only examples requiring at most three evidence chunks. We train on all remaining training data after the early stage.

The motivation is to stabilize early exploration. Very long contexts and many-hop evidence chains can lead to invalid actions and noisy rewards in the early stage. 

% The curriculum only controls the initial training stage; the full reward objective remains unchanged.

\subsubsection{Maximum Number of Actions}
To control computational cost, we set the action limits according to the empirical distribution of required evidence units in the training data. Specifically, we set the maximum final evidence memory size to $K_{\max}=6$, which covers all filtered training examples. We set the maximum numbers of add, drop, and link actions to: $N_{\mathrm{add}}=7, N_{\mathrm{drop}}=3, N_{\mathrm{link}}=4.$
These limits provide a small additional budget for exploratory evidence additions and subsequent removal of distractors, while preventing degenerate add--drop loops.

The action limits are explicitly stated in the rollout prompt. During training, we also enforce them in the parser. We parse actions sequentially from left to right and maintain the evidence memory online. If an action references a nonexistent evidence identifier, or exceeds the allowed action budget, we mark the action as invalid and do not assign it positive process reward. Over-budget actions are ignored for evidence-memory updates. 
% We do not retroactively remove earlier valid actions, since doing so would change the credit assignment of previous decisions. Therefore, when the evidence memory is full, a later add action is ignored unless the model first drops an existing evidence item.
For link actions, both referenced evidence identifiers must exist in the current evidence memory. For drop actions, the referenced evidence identifier must also exist in the current evidence memory. If the model generates a malformed action or an action with nonexistent identifiers, the action is ignored and receives no positive process reward.

\subsubsection{Computational Considerations}
\label{sec:computational_considerations}

\textsc{Maven} does not require training a separate process reward model. The verifier \(p_\psi\) is frozen and is used only during training to compute answer-token NLL under compact evidence memories. At inference time, the verifier is not used.

For each trajectory, the required verifier evaluations are:
\begin{itemize}
    \item the empty-memory value \(V_\psi(Q,\varnothing)\);
    \item evidence-state values for add and drop transitions;
    \item leave-one-out values \(V_\psi(Q,E_F\setminus\{e_i\})\) for final kept evidence;
    \item pair-removal values \(V_\psi(Q,E_F\setminus\{e_i,e_j\})\) for linked pairs.
\end{itemize}
With caps \(K_{\max}\) on final evidence size, \(N_{\max}\) on memory-edit actions, and \(L_{\max}\) on links, the number of verifier calls is bounded by:
\begin{equation}
    O(N_{\max} + K_{\max} + L_{\max}).
\end{equation}
These calls are inexpensive relative to long-context generation because the verifier conditions only on the question and selected evidence snippets, not on the full long context, and it scores the gold answer under teacher forcing rather than generating new text.

\subsubsection{Inference}
\label{sec:inference}

At inference time, the trained policy receives only the long context and question. It generates an editable evidence trajectory and a final answer:
\begin{equation}
    \hat\tau
    =
    \pi_\theta(C,Q).
\end{equation}
The answer inside the \texttt{<answer>} span is returned as the model prediction. 
% The intermediate evidence actions may optionally be retained as an interpretable rationale, but no verifier or reward computation is required during inference.

\subsection{Training Dataset}
\label{app:data}
We construct RL and SFT data from two sources: LongRLVR-style grounded long-context examples and public multi-hop QA datasets. Our goal is to build training examples that require evidence addition, synthesis, and revision, rather than examples that are merely long.

\subsubsection{LongRLVR Data}
\label{app:longrlvr_data}

We sample 3K examples from the LongRLVR training data. We retain examples satisfying the following conditions: (1) the context length is between 16K and 64K tokens; (2) the number of reference evidence chunks is between 2 and 5; (3) the original abstractive answer can be converted into a short and unambiguous answer list; (4) the example does not contain a single-evidence shortcut.

For the answer conversion step, we prompt Qwen3-32B (Thinking) to convert the original abstractive answer into a list of short answer aliases or keywords. We discard examples whose answer cannot be converted into a concise and unambiguous form. The resulting answer list is used for answer matching during training.

To remove examples with single-evidence shortcuts, we use the frozen verifier \(p_\psi\). Let \(G=\{g_1,\ldots,g_m\}\) be the reference evidence set. We discard an example if one evidence chunk alone accounts for most of the value of the full evidence set:
\begin{equation}
    \max_{g_i\in G}
    V_\psi(Q,\{g_i\})
    >
    0.8\,V_\psi(Q,G).
\end{equation}
This filter encourages examples where multiple evidence chunks are needed jointly.

\subsubsection{Multi-Hop QA Data}
\label{app:multihop_data}
We construct 6K RL examples from public multi-hop QA datasets, including 1K examples from HotpotQA, 1K examples from 2WikiMultiHopQA, and 4K examples from MuSiQue. We run Qwen3-30B-A3B-Instruct-2507 eight times and remove examples with pass rate 0 or 1. We also apply the same single-evidence shortcut filter used for LongRLVR-derived data.

We construct long contexts from 8K to 64K by combining gold supporting evidence with distractors. Distractors come from (1) randomly sampled documents from filtered-out examples in the original datasets; (2) teacher-generated hard distractors produced by Qwen3-235B-A22B-Thinking-2507~\citep{yang2025qwen3}.
For hard distractors, we prompt the teacher model to generate passages that are lexically or semantically related to the question but do not support the gold answer. In particular, we construct partial-chain distractors that share entities, relations, or answer types with the gold reasoning chain but lead to an incorrect conclusion. 
We discard distractors that contain the gold answer, or make the question ambiguous.
% Gold evidence chunks are distributed across different positions in the context to reduce positional shortcuts. Each context unit is assigned a source identifier, which is used by the model in \texttt{<add>} actions.

\subsubsection{Cold-Start SFT Data}
\label{app:sft_data}
We construct 2K cold-start SFT examples from the same source datasets. The context length of SFT examples ranges from 4K to 16K tokens. We first ensure that Qwen3-235B-A22B-Thinking-2507 can answer each selected example correctly. We then prompt it to produce an editable evidence-memory trajectory using \texttt{add}, \texttt{link}, \texttt{drop}, and \texttt{answer} actions.

Most SFT trajectories contain only useful evidence additions and links. To teach the model how to revise its evidence memory, we include drop actions in at most 20\% of the SFT examples. For these examples, the teacher is instructed to intentionally add one plausible distractor evidence item and then drop it after identifying that it does not support the correct reasoning chain. 
% We retain only trajectories that satisfy all of the following conditions:
% \begin{enumerate}
%     \item the final answer is correct;
%     \item all source identifiers are valid;
%     \item all drop and link actions refer to existing evidence identifiers;
%     \item the trajectory is parseable under our action grammar;
%     \item the trajectory does not exceed the action limits.
% \end{enumerate}

During RL, trajectories are generated online by the policy using the rollout prompt in Appendix~\ref{app:prompts}.

\subsection{Implementation and Evaluation Details}
\label{app:implementation}
\subsubsection{Implementation Details}
We train three policy models: Llama-3.1-8B-Instruct (8.01B parameters), Qwen2.5-14B-Instruct (14.7B parameters), and Qwen3-30B-A3B-Instruct-2507 (30.5B total parameters, 3.3B activated). We use Qwen3-4B-Instruct-2507 as the frozen verifier.
For cold-start SFT, we use a learning rate of \(2\times10^{-5}\), a global batch size of 16, and 30 warmup steps. We train Llama-3.1-8B-Instruct and Qwen2.5-14B-Instruct for 2 epochs, and Qwen3-30B-A3B-Instruct-2507 for 1 epoch.
For RL training, we use GRPO with rollout group size \(G=8\) and global prompt batch size 16. We train for one epoch over the RL training set. We use a learning rate of \(1\times10^{-6}\), 10 warmup steps, and cosine learning-rate decay. The maximum response length is 4096 tokens. The KL coefficient is set to \(0.001\), and the GRPO clipping threshold is set to \(0.2\).
For rollouts, we sample with temperature \(0.8\) and \texttt{top\_p=0.95}.
For reward computation, we set the local reward clipping threshold to \(c=0.3\). The reward weights for add, drop, link, and answer rewards are set to: $\lambda_{\mathrm{add}}=0.5, \lambda_{\mathrm{drop}}=0.2, \lambda_{\mathrm{link}}=0.3, \lambda_{\mathrm{ans}}=1.0.$ For the add reward, we set \(\alpha=0.6\).

We implement RL training using \texttt{verl}. The frozen verifier is used only during training to compute teacher-forced answer NLL over selected evidence memories.

Experiments are conducted on 8 NVIDIA A100 GPUs and 4 AMD MI250X GPUs. The total training budget for SFT and RL training across three models, averaged on two hardware environments is approximately 1,250 GPU hours, computed as the number of GPUs multiplied by wall-clock training time.

\subsubsection{Evaluation Details}
\label{app:evaluation}
We use vLLM for benchmark inference and follow the official inference configuration of each benchmark whenever available. We report the average accuracy across three runs under the official evaluation protocol. For RULER, we evaluate selected subsets: Needle-in-a-Haystack (NIAH), variable tracking, and SQuAD QA subsets, and evaluate them at 64K and 128K context lengths. 

We construct a held-out diagnostic set of 150 examples from HotpotQA, 2WikiMultiHopQA, and MuSiQue using the same filtering and distractor-construction strategy as the training data. These examples are not used during SFT or RL training. Each diagnostic example contains gold evidence annotations and distractor labels, allowing us to evaluate evidence-process behavior.

\subsection{Prompts}
\label{app:prompts}

\subsubsection{RL Rollout Prompt}
\label{app:rollout_prompt}

The following prompt is used for RL rollouts.

\begin{tcolorbox}[breakable, colback=gray!5, colframe=gray!40, title=RL Rollout Prompt]
You are given a long context and a question. Your task is to answer the question.

You may use the following actions:

\texttt{<add id=i>} \\
Add one useful evidence unit from the context to your evidence memory. The field \texttt{id=i} must refer to a valid source identifier in the context. Briefly summarize the evidence and explain briefly how evidence \texttt{i} is relevant to the question.\\
\texttt{</add>}

\texttt{<link ids=i,j>} \\
Explain how evidence \texttt{i} and evidence \texttt{j} jointly support the answer. Use this when two evidence pieces need to be combined. \\
\texttt{</link>}

\texttt{<drop id=i>} \\
Remove evidence \texttt{i} if it is irrelevant, redundant, or misleading. Explain briefly why it should be removed. \\
\texttt{</drop>}

\texttt{<answer>} \\
Give the final answer (without extra text) only after you have collected enough evidence. \\
\texttt{</answer>}

Rules:
\begin{itemize}
    \item Use only source identifiers that appear in the context.
    \item Do not invent evidence.
    \item A link or drop action can only refer to an evidence id that has already been added and not dropped.
    \item You may use at most 7 add actions, 3 drop actions, and 4 link actions.
    \item Stop after producing the final \texttt{<answer>} action.
\end{itemize}

Long context:
\{\{CONTEXT\}\}

Question:
\{\{QUESTION\}\}

Now produce your trajectory using the allowed actions.
\end{tcolorbox}

\subsubsection{SFT Trajectory Generation Prompt}
\label{app:sft_prompt}

We use the following prompt to generate cold-start SFT trajectories.

\begin{tcolorbox}[breakable, colback=gray!5, colframe=gray!40, title=SFT Trajectory Generation Prompt]
You are an expert annotator creating a teacher trajectory used to train a long-context QA model.

You are given:
\begin{itemize}
    \item a long context segmented into source units;
    \item a question;
    \item the gold answer;
    \item the gold supporting source identifiers;
    \item optional hard distractor source identifiers.
\end{itemize}

Your task is to write an ideal editable evidence-memory trajectory to reason over the question using the following actions:
\texttt{<add>}, \texttt{<link>}, \texttt{<drop>}, and \texttt{<answer>}.

Here is the required format for each action:
\texttt{<add id=i>} \\
Add one gold evidence unit from the context to your evidence memory. Briefly summarize the evidence and explain briefly how evidence \texttt{i} is relevant to the question.\\
\texttt{</add>}

\texttt{<link ids=i,j>} \\
Explain how evidence \texttt{i} and evidence \texttt{j} jointly support the answer. Use this when two evidence pieces need to be combined. \\
\texttt{</link>}

\texttt{<drop id=i>} \\
Remove distractor \texttt{i} if it is irrelevant, redundant, or misleading. Explain briefly why it should be removed. \\
\texttt{</drop>}

\texttt{<answer>} \\
Give the final answer. \\
\texttt{</answer>}

Requirements:
\begin{itemize}
    \item The final answer must exactly match or be equivalent to the gold answer.
    \item Add the necessary gold evidence using valid source identifiers.
    \item Use link actions to connect complementary evidence pieces.
    \item Keep explanations concise.
    \item Do not exceed 7 add actions, 3 drop actions, or 4 link actions.
\end{itemize}

If \texttt{DROP\_EXAMPLE = true}, intentionally add one plausible distractor evidence item first, then later drop it after explaining why it does not support the correct reasoning chain. If \texttt{DROP\_EXAMPLE = false}, do not add unnecessary distractors.

Context:
\{\{CONTEXT\}\}

Question:
\{\{QUESTION\}\}

Gold answer:
\{\{ANSWER\}\}

Gold supporting source ids:
\{\{GOLD\_SOURCES\}\}

Optional hard distractor source ids:
\{\{DISTRACTOR\_SOURCES\}\}

DROP\_EXAMPLE:
\{\{DROP\_EXAMPLE\}\}

Now write the trajectory.
\end{tcolorbox}

\subsection{Ablation (cont.)}
\label{app:ablation}

\begin{figure*}[t]
    \centering
    \includegraphics[width=0.95\textwidth]{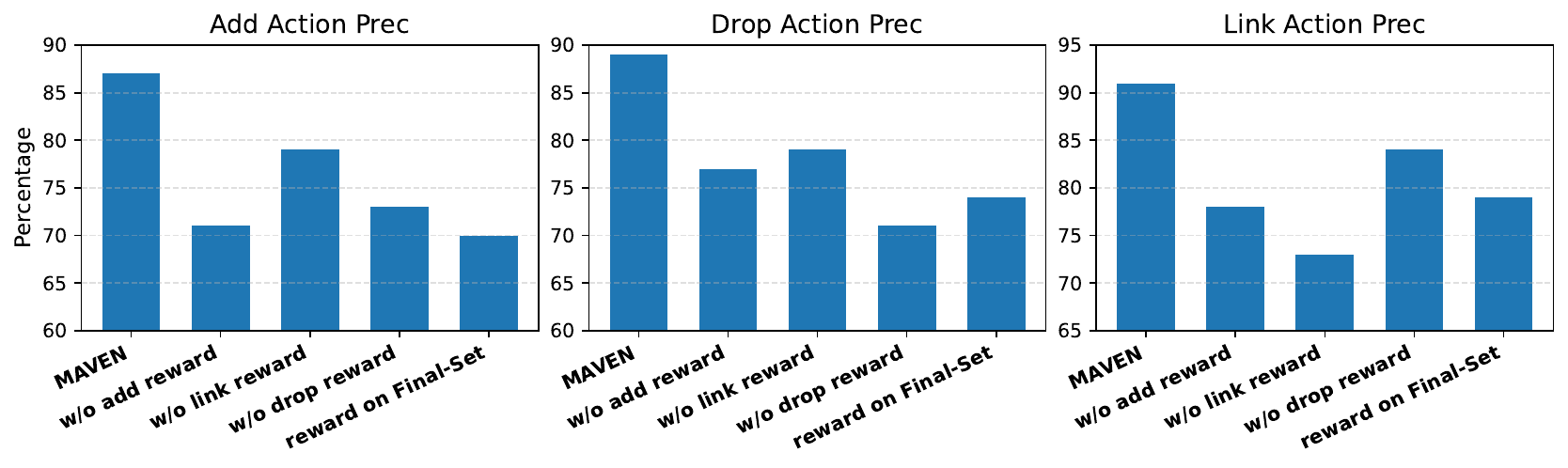}
    \caption{Impact of action rewards. We measure Add/Link/Drop Action Precision.}
    \label{fig:ablation2}
\end{figure*}

Figure~\ref{fig:ablation2} presents more ablation results on action rewards. We measure the precision of add/drop/link actions, which is defined as: (1) whether an evidence chunk selected by an add action is a gold evidence chunk; (2) whether an evidence chunk dropped by a drop action is not from the gold evidence set; (3) whether two evidence chunks linked by a link action are truly relevant. Before evaluating the link action, we used Gemini 3 to construct a reference link set for this diagnostic set. 
Based on the result, we draw the same conclusion as we illustrated in Sec.~\ref{sec:experiment_action_rewards}.

\subsection{Choice of Verifier}
\begin{table}[t]
\centering
\small
\caption{Effect of verifier choice on LongBench v2. We observe only minor differences between Qwen3-4B and Qwen2.5-7B verifiers.}
\label{tab:verifier_results}
\resizebox{0.95\linewidth}{!}{\begin{tabular}{lcccc
}
\toprule
\multirow{2}{*}{\textbf{Model}} &  \multicolumn{4}{c}{\textbf{LongBench v2}}  \\
\cmidrule(lr){2-5} 
 & {\textbf{Short}} & {\textbf{Medium}} & {\textbf{Long}} & {\textbf{Overall}} \\
\midrule
LLaMA-3.1-8B &&&&\\
\quad w/ Qwen3-4B &39.8&36.2&32.1&36.6 \\
\quad w/  Qwen2.5-7B &40.2&35.8&32.3&36.6 \\
\midrule
Qwen2.5-14B &&&&\\
\quad w/ Qwen3-4B &51.5&40.2&37.0&43.5 \\
\quad w/  Qwen2.5-7B &51.8&40.4&36.9& 43.7 \\
\bottomrule
\end{tabular}
}
\end{table}
Table~\ref{tab:verifier_results} shows results with two frozen verifiers.

\subsection{Hyperparameter Analysis}

\label{app:hyperparameter}
\begin{figure}[t]
    \centering
    \includegraphics[width=1.0\linewidth]{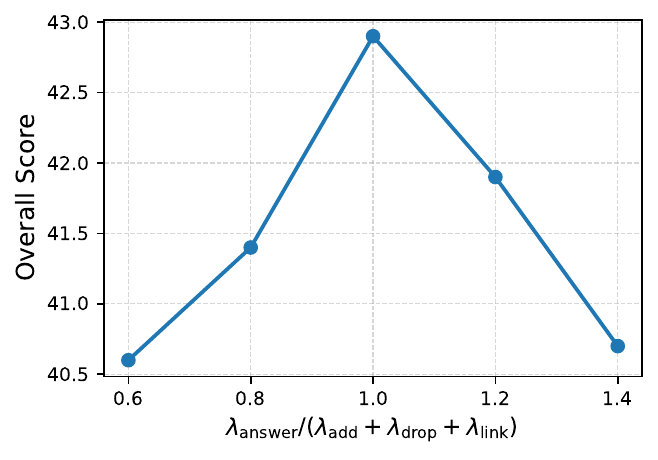}
    \caption{Overall score on the subset of LongBench v2 from Qwen2.5-14B. Vary the ratio between the answer reward weight and the sum of process reward weights.
    }
    \label{fig:hp_answer}
\end{figure}

\begin{figure*}[t]
    \centering
    \includegraphics[width=0.95\textwidth]{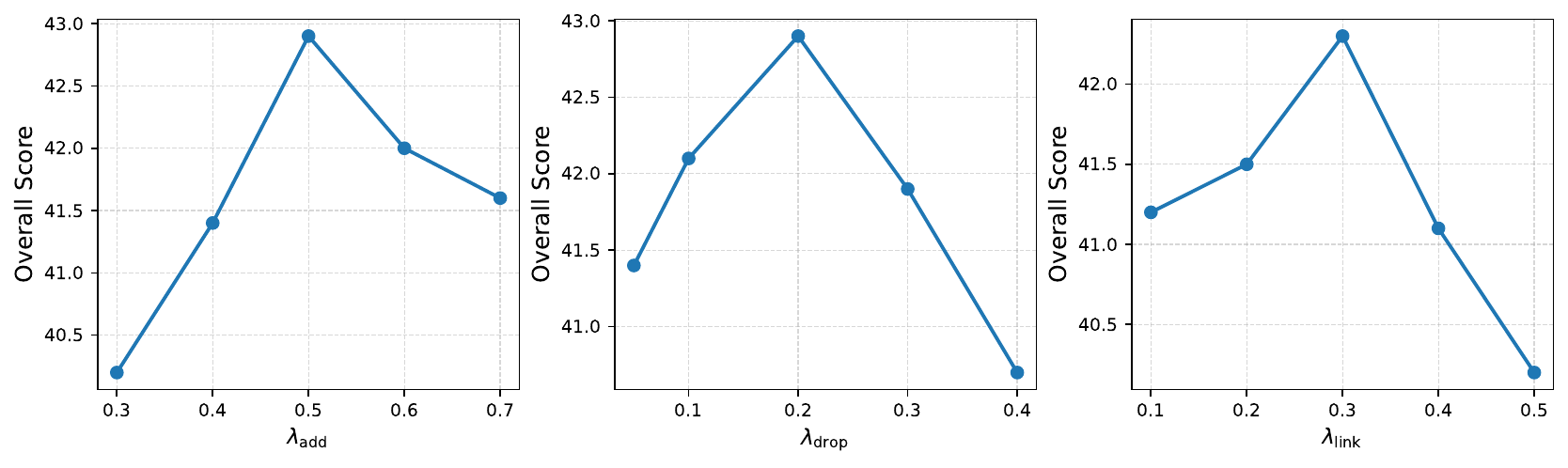}
    \caption{Overall score on the subset of LongBench v2 from Qwen2.5-14B. We vary the add, drop, and link reward weights.}
    \label{fig:hp}
\end{figure*}

\begin{figure*}[t]
    \centering
    \includegraphics[width=0.95\textwidth]{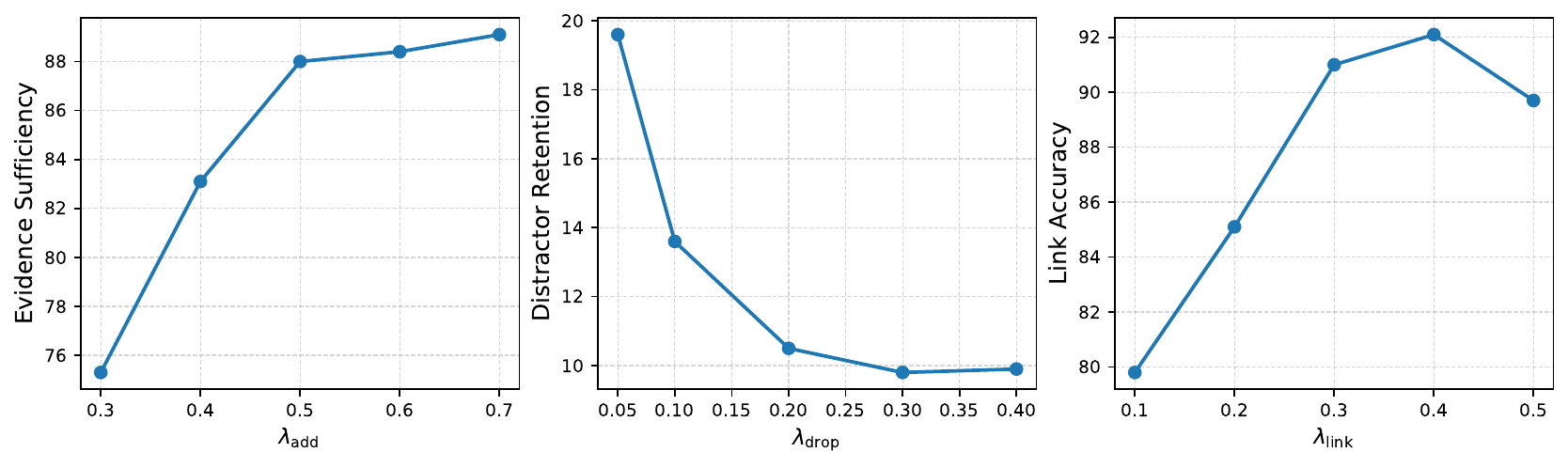}
    \caption{Corresponding diagnostic metrics from Qwen2.5-14B. We vary the weight value. Larger add weight improves evidence sufficiency, larger drop weight reduces distractor retention, and moderate link weight improves link precision.}
    \label{fig:hp2}
\end{figure*}

\begin{figure}[t]
    \centering
    \includegraphics[width=1.0\linewidth]{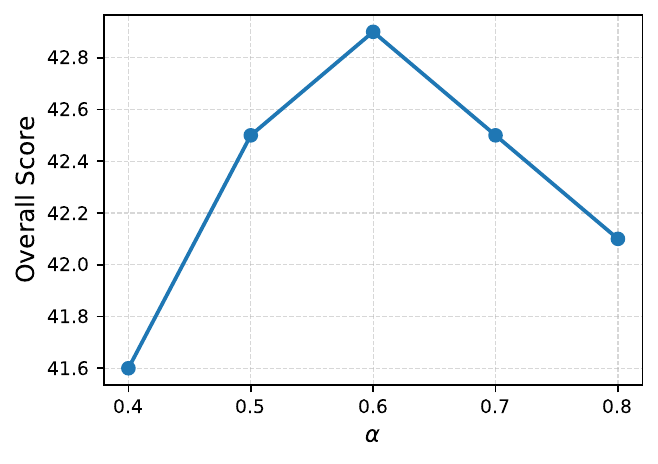}
    \caption{Overall score on the subset of LongBench v2 from Qwen2.5-14B. We vary \(\alpha\), which balances online CAIG and hindsight credit in the add reward.
    }
    \label{fig:hp_alpha}
\end{figure}
We analyze the sensitivity of \textsc{Maven} to the relative weighting of reward components by an ablation study on a fixed subset of 120 LongBench v2.
\paragraph{Answer-process balance.}
We first vary the ratio between the answer reward weight and the sum of process reward weights:$
    \frac{
        \lambda_{\mathrm{ans}}
    }{
        \lambda_{\mathrm{add}}
        +
        \lambda_{\mathrm{drop}}
        +
        \lambda_{\mathrm{link}}
    }$.

We fix the sum of all reward weights to 2 and keep the relative ratio among process rewards fixed to the default proportion: $\lambda_{\mathrm{add}}:\lambda_{\mathrm{drop}}:\lambda_{\mathrm{link}} = 5:2:3$.

As shown in Figure~\ref{fig:hp_answer}, performance peaks around \(r_{\mathrm{ans}}=1.0\), which corresponds to a balanced weighting between final answer correctness and process-level evidence supervision. Smaller ratios underweight the answer reward, while larger ratios reduce the influence of process rewards.

\paragraph{Process reward weights.}
We then study the relative weights among add, drop, and link rewards. In these experiments, we fix $  \lambda_{\mathrm{ans}}=1.0, \lambda_{\mathrm{add}}+\lambda_{\mathrm{drop}}+\lambda_{\mathrm{link}}=1.0$.
    
When varying one process weight, we distribute the remaining process weight to the other two actions according to their default ratio. For example, when varying \({\lambda}_{\mathrm{add}}\), the remaining process weight is divided between drop and link rewards with ratio \(2:3\).

Figure~\ref{fig:hp} and \ref{fig:hp2} shows that the best overall performance is achieved near the default process ratio: ${\lambda}_{\mathrm{add}}=0.5,{\lambda}_{\mathrm{drop}}=0.2,{\lambda}_{\mathrm{link}}=0.3$.
Increasing the add weight improves evidence sufficiency, but overly large add weight does not further improve final accuracy. Increasing the drop weight reduces distractor retention, while too much drop reward may encourage the model to remove useful evidence. Link precision improves as the link weight increases to a moderate value, but excessive link reward can hurt overall answer performance.

\paragraph{Online versus hindsight add credit.}
Finally, we vary \(\alpha\) in the add reward.
As shown in Figure~\ref{fig:hp_alpha}, \(\alpha=0.6\) gives the best overall performance. This suggests that online marginal progress should receive slightly more weight than hindsight credit, while both terms are necessary for stable evidence construction.

\subsection{Performance on Short Reasoning Tasks}
 \begin{table}
    \centering
    \small
    \caption{Performance on MMLU-Pro. The model is evaluated by generating CoT.} 
    \label{tab:mmlu}
    \resizebox{0.95\linewidth}{!}{
    \begin{tabular}{lcc} 
        \toprule
      Model  & Base & \textsc{MAVEN}   \\
        \midrule
    Llama-3.1-8B-Instruct  &44.3& 45.7\\            
     Qwen2.5-14B-Instruct & 64.0 & 64.2 \\
     Qwen3-30B-A3B-Instruct-2507 &77.5  & 76.9 \\
  
        \bottomrule
    \end{tabular}
    }
\end{table}
We evaluate whether training with \textsc{Maven} affects general short-context reasoning ability. We report results on MMLU-Pro~\citep{wang2024mmlu}, using CoT generation during evaluation. As shown in Table~\ref{tab:mmlu}, \textsc{Maven} does not cause clear degradation.

\section{Potential Risks}
MAVEN improves the ability of LLMs to navigate, revise, and synthesize evidence from long contexts. While this can benefit document understanding, question answering, and research assistance, stronger long-context reasoning may also increase the effectiveness of harmful applications that require processing large volumes of information. In addition, the generated evidence trajectories should not be treated as guaranteed faithful explanations. Our experiments are conducted in a research setting, and deployment in high-stakes domains should require additional human oversight, stronger faithfulness checks, and domain-specific safety evaluation.

\section{LLM Usage}
We used large language models to assist with writing, grammar improvement, and clarity edits. We also used teacher LLMs to generate hard distractors and cold-start SFT trajectories, as described in the data construction appendix. All model-generated text used in the paper was reviewed and edited by the authors. LLMs were not used to make final scientific claims or to replace author judgment in experimental analysis.

\section{Artifact Use}

We use existing datasets and models only for research purposes. The training and evaluation data are derived from public research artifacts, including HotpotQA, 2WikiMultiHopQA, MuSiQue, LongRLVR data, LongBench v2, LongReason, and RULER. We follow the licenses and terms of the original artifacts. HotpotQA is distributed under CC BY-SA 4.0; 2WikiMultiHopQA is distributed under Apache 2.0; and MuSiQue is distributed under CC BY 4.0. We use Llama models under the Llama 3.1 Community License and Qwen models under their released model licenses. Any derivative data released from this work should preserve the usage restrictions and attribution requirements of the source artifacts and should be used only for research purposes.

We do not collect data from human participants. Our data are derived from public QA datasets. Since some QA examples may mention public entities or persons as part of the question-answering task, we do not anonymize entity names that are necessary for answering.

\end{document}